\documentclass[runningheads]{llncs}

\usepackage[final,year=2026]{eccv}
\usepackage[breaklinks,colorlinks]{hyperref}
\usepackage{eccvabbrv}
\usepackage{graphicx}
\usepackage{booktabs}
\usepackage{multirow}
\usepackage[table]{xcolor} 
\usepackage{algorithm}
\usepackage{algpseudocode}
\usepackage{amsmath,amssymb}
\usepackage[accsupp]{axessibility}

\usepackage{enumitem}

\begin{document}

\title{TRaM-VSR: Importance-Aware Token Routing and Merging for One-Step Diffusion Video Super-Resolution}
\titlerunning{TRaM-VSR}

\author{
Sicheng Gao\inst{1,2}
\and Zhuyun Zhou\inst{2}
\and Yixuan Liu\inst{1}
\and Tong Shen\inst{1}
\and 
\\
Zongwei Wu\inst{2}\thanks{Corresponding author.}
\and Radu Timofte\inst{2}
}
\authorrunning{S. Gao et al.}
\institute{
Advanced Micro Devices Inc. \and Computer Vision Lab, CAIDAS \& IFI, University of W\"urzburg
}

\maketitle

\begin{abstract}
Video super-resolution (VSR) using large-scale Diffusion Transformer (DiT) priors achieves exceptional perceptual quality but is often impractical due to the quadratic computational cost of processing dense spatio-temporal token sequences. Existing efficiency-oriented methods risk irreversible detail loss and temporal flickering, a vulnerability especially pronounced in one-step diffusion models. To address this, we propose TRaM-VSR, a Token Routing and Merging framework for adaptive token allocation, leveraging both context-aware video priors and network-level priors. First, token importance is estimated by fusing motion-sensitive temporal cues with semantic text similarity, isolating dynamic objects and structural boundaries. Next, this importance is further calibrated and adjusted by an offline planner to guide routing across optimally grouped network blocks. Technically, within each routed group, structurally critical tokens are processed in a high-fidelity local stream, while less informative tokens are aggregated into a compact global stream, both modulated by network depth and aligned with the multigranular nature of diffusion models. Extensive experiments show that TRaM-VSR accelerates inference significantly while preserving state-of-the-art reconstruction quality and robust temporal consistency. The code is available at \url{https://github.com/Ree1s/TRaM-VSR}.
\end{abstract}

\section{Introduction}
\label{sec:intro}

Video super-resolution (VSR) aims to recover high-resolution (HR) videos from low-quality (LQ) inputs while preserving temporal consistency across frames. Recently, diffusion-based approaches have achieved remarkable progress, largely driven by Diffusion Transformers (DiTs) and large-scale text-to-video priors~\cite{peebles2023dit,blattmann2023align,esser2023structure,singer2023make,villegas2023phenaki,khachatryan2023text2video}. Leveraging powerful generative priors, recent VSR works~\cite{zhou2024upscale,wang2025seedvr,he2024venhancer} achieve impressive performance in super-resolution. However, these models incur substantial computational cost, as dense spatio-temporal tokens must be processed through deep transformer stacks with quadratic attention complexity.

To improve efficiency, a line of research explores lightweight generation pipelines, among which one-step diffusion has emerged as a promising paradigm~\cite{chen2025dove,sun2025onestepvsr,li2025osdiffvsr}. By collapsing the iterative denoising process into a single forward pass, these approaches substantially reduce inference latency. However, this efficiency often comes at the cost of temporal stability~\cite{chen2025dove,sun2025onestepvsr}. Without the progressive refinement provided by multiple diffusion steps, generation errors cannot be gradually corrected and may accumulate over time, leading to artifacts such as shadow flickering and temporally inconsistent textures across frames. This issue becomes particularly pronounced when spatio-temporal dependencies are not sufficiently modeled within the one-step diffusion process.

\begin{figure}[t]
  \centering
  \includegraphics[width=\linewidth]{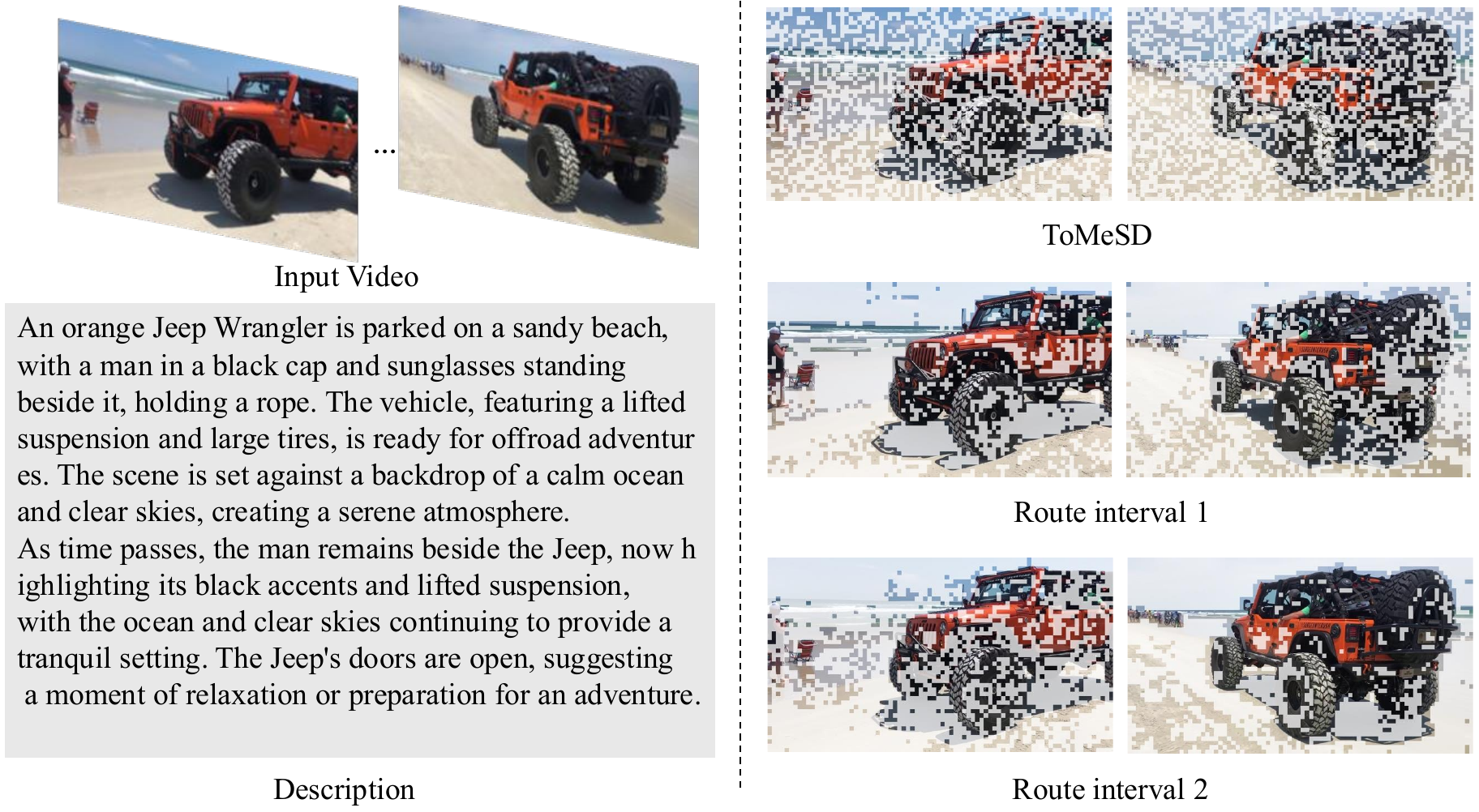}
  \caption{
    \textbf{Motivation of TRaM-VSR.} ToMeSD~\cite{bolya2023tomesd} (top right) yields scattered heatmaps that miss dynamic objects and prompt semantics. Instead, TRaM-VSR exploits the coarse-to-fine nature of DiTs: our offline planner locates safe compressible depths (Intervals 1 and 2), where our semantic-temporal scoring (bottom right) precisely highlights critical tokens. Notably, this scoring forms complementary focus regions across intervals, synergistically preserving both global motion and fine details. More discussion and analysis can be found in Section \ref{discussion}.
  }
  \label{fig:introduction}
\end{figure}

However, effectively modeling spatio-temporal cues is challenging, as video frames can be affected by complex motion patterns arising from both camera ego-motion and independently moving objects. When the learning model has limited capacity, as in the case of one-step diffusion, uniformly processing all tokens becomes difficult and can introduce significant errors. To improve efficiency, some studies~\cite{bolya2023tome,li2024vidtome,dong2025tinysr,fang2025tinyfusion} have explored token merging or token dropping strategies. However, these accelerations are typically applied in a uniform manner \cite{castells2024ldpruner,bolya2023tomesd}, as shown in Figure~\ref{fig:introduction}. Consequently, the modeling capacity is not allocated where it is most needed, leading to degraded temporal coherence. This observation motivates us to introduce a more context-aware perspective. 

In this paper, we propose \textbf{TRaM-VSR}, a \emph{Token Routing and Merging} framework for efficient video super-resolution. The core idea is to perform adaptive token allocation by jointly leveraging context-aware \textit{video priors} and \textit{network-level priors}. For the video context, we learn a token importance map that captures both spatial and temporal relationships across frames, allowing the model to rank tokens according to their relevance for high-fidelity reconstruction. This selection is further guided by input text prompts, ensuring semantic alignment. Consequently, tokens corresponding to dynamic objects, structural boundaries, and temporally informative regions should receive higher importance, while less informative regions can be processed with reduced computational effort.

We then correlate the video-derived token importance with depth-aware contextual cues from the backbone network. In the diffusion-based architecture~\cite{peebles2023dit}, each block of the DiT backbone functions as an independent denoising operator, exhibiting distinct behaviors at different depths. Inspired by complex systems~\cite{tang2023emergent}, we hypothesize that organizing these operators at multiple granularities can induce emergent modeling capabilities that surpass the capabilities of individual blocks. Accordingly, we group DiT blocks into sets of varying granularities, with each group producing an independent dropping score that serves as a calibration prior, dependent solely on the compositional structure and depth of the network. By combining the video-derived token importance with the corresponding network-level dropping prior, our framework achieves holistic adaptive token routing and dropping throughout the transformer hierarchy. Extensive evaluation on VSR benchmarks shows that our method can achieve high-fidelity output at a very efficient scale. Our contributions are summarized as follows:

\begin{itemize}[noitemsep, topsep=0pt]
    \item We introduce a novel semantic--temporal importance scoring mechanism. Compared to generic compression metrics, our method effectively isolates motion-critical and prompt-relevant tokens, avoiding the temporal flickering induced by motion-blind pruning.
    \item We explicitly leverage the coarse-to-fine depth dynamics of DiTs to design a deterministic offline planner, precisely localizing optimal routing intervals to prevent catastrophic detail loss in one-step VSR.
    \item We design a highly efficient two-stream token pathway featuring explicit identity restoration. Extensive experiments demonstrate that TRaM-VSR significantly accelerates inference while preserving state-of-the-art quality and temporal consistency.
\end{itemize}

\section{Related Work}
\label{sec:related}

\noindent \textbf{Diffusion Priors for Video Restoration}:
Diffusion models, particularly those leveraging large-scale text-to-video (T2V) priors, have recently dominated video super-resolution (VSR) due to their robust perceptual quality and highly flexible conditioning~\cite{wang2024lavie,peebles2023dit}. Systems such as Upscale-A-Video~\cite{zhou2024upscale}, SeedVR~\cite{wang2025seedvr}, and VEnhancer~\cite{he2024venhancer} demonstrate how adapting these rich spatio-temporal priors significantly improves real-world restoration by handling complex motion and severe degradations~\cite{wu2024seesr,yang2024motion,xie2025star}. To address the inherent latency of iterative sampling, recent advancements have introduced one-step diffusion VSR models, notably DOVE~\cite{chen2025dove}, One-Step VSR~\cite{sun2025onestepvsr}, OS-DiffVSR~\cite{li2025osdiffvsr}, and UltraVSR~\cite{liu2025ultravsr}. Despite drastically reducing the number of sampling steps, these architectures still inherit the massive spatial and temporal token footprints of their foundational DiT backbones. Consequently, shifting the efficiency bottleneck from temporal iteration to spatial token computation remains an open challenge.

\noindent \textbf{Efficiency in Diffusion Models}:
Improving the inference efficiency of diffusion models has inspired numerous architectural and system-level accelerations. Traditional efforts primarily target sampling dynamics, reducing the number of denoising iterations through advanced ODE/SDE solvers~\cite{dockhorn2022genie,karras2024analyzing} or distillation techniques like Latent Consistency Models (LCM)~\cite{luo2023lcm}. Other orthogonal directions include attention scheduling~\cite{wang2024attentiondriven} and hardware-aware system optimizations~\cite{shen2025efficient}. While these methods effectively accelerate multi-step models, they are fundamentally complementary to token-level routing. For one-step VSR frameworks, where sampling-step reduction is already saturated, alleviating the internal token complexity is not trivial.

\noindent \textbf{Token Compression in DiT Backbones}:
Token compression directly reduces sequence length to mitigate the computational overhead of self-attention. Methods like Token Merging (ToMe)~\cite{bolya2023tome,bolya2023tomesd} and its video extension, VVidToMe~\cite{li2024vidtome}, aggregate redundant tokens, while various pruning adaptations selectively discard them to improve diffusion efficiency~\cite{dong2025tinysr,castells2024ldpruner,fang2025tinyfusion,fang2023structural}. However, applying naive token compression to DiT-based VSR often degrades high-frequency textures and temporal consistency, particularly when token selection ignores motion dynamics or is applied uniformly across all network depths. Unlike existing literature that relies on static or depth-agnostic compression, TRaM-VSR introduces a dynamic, importance-aware routing mechanism. By coupling risk-guided interval planning with a motion-sensitive TCG saliency cue, our method strictly preserves the generative priors of DiT backbones while maximizing inference throughput. Importantly, TCG is exclusively utilized for routing and does not alter the underlying denoising trajectory.

\section{Preliminaries}
\label{sec:preliminaries}

 We operate in the latent space of a pretrained Variational Autoencoder (VAE) to process a low-quality (LQ) video $\mathbf{x}^{\mathrm{lq}} \in \mathbb{R}^{B\times 3\times F\times H\times W}$ and its high-quality (HQ) counterpart $\mathbf{x}^{\mathrm{hq}}$. The VAE encoder $\mathcal{E}$ maps these videos to their respective latent representations:
\begin{equation}
\mathbf{z}^{\mathrm{lq}}=\mathcal{E}(\mathbf{x}^{\mathrm{lq}}),\qquad
\mathbf{z}^{\mathrm{hq}}=\mathcal{E}(\mathbf{x}^{\mathrm{hq}}).
\end{equation}

Following the standard latent diffusion formulation, a clean latent $\mathbf{z}_0$ is perturbed by Gaussian noise. At a fixed super-resolution timestep $t$, the noised latent is
\begin{equation}
\mathbf{z}_{t}=\sqrt{\bar{\alpha}_{t}}\,\mathbf{z}_{0}
+\sqrt{1-\bar{\alpha}_{t}}\,\boldsymbol{\epsilon},\qquad
\boldsymbol{\epsilon}\sim\mathcal{N}(\mathbf{0},\mathbf{I}),
\label{eq:prelim_forward}
\end{equation}
where $\bar{\alpha}_t\in(0,1)$ is the cumulative noise level determined by the noise schedule.

 Recent one-step VSR frameworks~\cite{chen2025dove,sun2025onestepvsr,li2025osdiffvsr} train the denoiser to predict the velocity target $\mathbf{v}_{\theta}$. Conditioned on the noised latent $\mathbf{z}_{t}$, a text prompt $\mathbf{c}$, and the timestep $t$, the network directly estimates the clean latent in a single forward pass:
\begin{equation}
\hat{\mathbf{z}}_{0}
=
\sqrt{\bar{\alpha}_{t}}\,\mathbf{z}_{t}
-
\sqrt{1-\bar{\alpha}_{t}}\,\mathbf{v}_{\theta}(\mathbf{z}_{t},\mathbf{c},t).
\label{eq:prelim_denoise}
\end{equation}
The final restored video is then decoded via the VAE decoder as $\hat{\mathbf{x}}=\mathcal{D}(\hat{\mathbf{z}}_{0})$.

\section{Methodology}
\label{sec:method}
\subsection{Overview}

\label{sec:method_overview}

\begin{figure}[t]
  \centering
  \includegraphics[width=\linewidth]{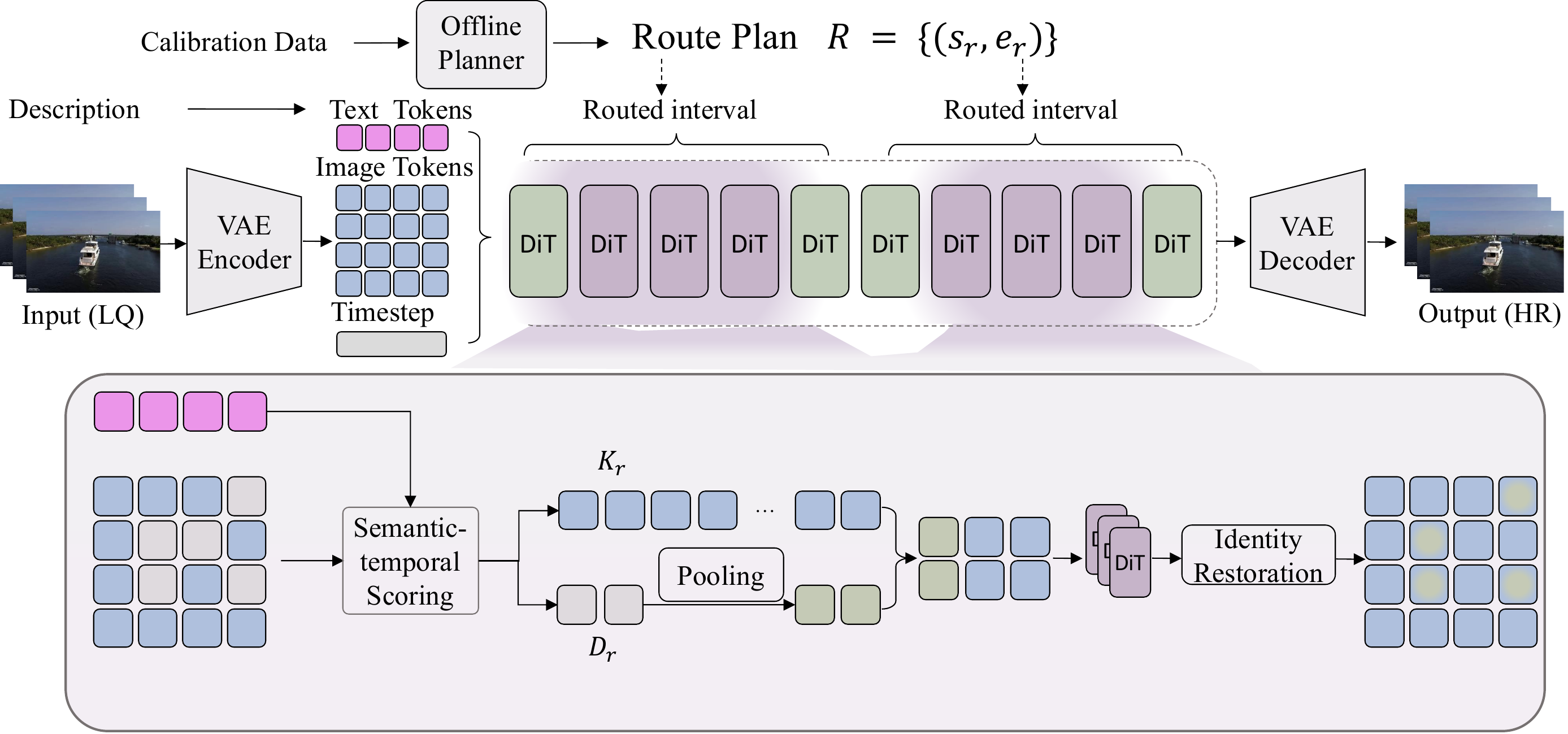}
  \caption{Overview of the proposed TRaM-VSR pipeline. Video frames are first analyzed to estimate token importance. This video-aware prior is then combined with network-level priors, organizing tokens across multiple block granularities.}
  \label{fig:pipeline}
\end{figure}

Our goal is to reduce the token-computation cost of a one-step DiT denoiser without sacrificing restoration quality. As illustrated in Figure~\ref{fig:pipeline}, TRaM-VSR confines token compaction to explicitly selected depth intervals and maintains full-resolution processing elsewhere. Each routed interval is specified by its start layer and end layer; routing begins at the start layer, compact processing is applied within the interval, and full-resolution tokens are seamlessly restored at the end layer.

TRaM-VSR comprises three core components: (1) risk-guided offline planning for route intervals (Sec.~\ref{sec:route_plan}), (2) importance-aware token routing at the route entry (Sec.~\ref{sec:token_importance}), and (3) a two-stream route-and-merge process inside routed intervals with identity restoration at the route exit (Sec.~\ref{sec:two_stream}). To reduce train--test mismatch, we follow a standard two-stage one-step training protocol and keep the routing mechanism enabled during training~\cite{chen2025dove,sun2025onestepvsr,li2025osdiffvsr}.
\subsection{Risk-Guided Route Interval Planning}
\label{sec:route_plan}

This stage determines a small set of non-overlapping routed intervals, each specified by its start layer $s_r$ and end layer $e_r$.
The token drop ratio is treated as a fixed hyper-parameter $\rho$ shared across intervals.

An offline analysis on the uncompressed baseline model computes two layer-wise signals:
a local update magnitude $\Delta_\ell$ and an end-to-end perturbation shift $S_\ell$.
Both signals are normalized by their layer-wise means to enable cross-depth comparison.
The perturbation shift $S_\ell$ is measured as the deviation in the final restored output when routing is simulated exclusively at layer $\ell$ (averaged over a small calibration set), which captures how local perturbations propagate to the final prediction.

With these signals, the routing risk proxy is defined as
\begin{equation}
rss_\ell=\hat{\Delta}_\ell\cdot\hat{S}_\ell,
\label{eq:rss}
\end{equation}
and mapped to a smooth routing gate
\begin{equation}
gate_\ell=\sigma\!\left(\frac{\tau-rss_\ell}{s}\right),
\qquad gate_\ell\in(0,1),
\label{eq:gate}
\end{equation}
where $\sigma(\cdot)$ is the sigmoid, $\tau$ is a risk threshold, and $s$ controls smoothness.
Lower-risk layers yield larger $gate_\ell$.

We score each candidate window $\Omega=[s,s+W-1]$ by its average gate:
\begin{equation}
\mathrm{Score}(\Omega)=\frac{1}{W}\sum_{\ell\in \Omega}\mathrm{gate}_\ell.
\label{eq:window_score}
\end{equation}
We then greedily select $R$ non-overlapping windows in descending order of $\mathrm{Score}(\Omega)$,
\looseness=-1
\subsection{Importance-Aware Token Routing}
\label{sec:token_importance}

Given a planned interval $(s_r,e_r)$ from Sec.~\ref{sec:route_plan}, the routing mechanism dictates which image tokens to preserve at the route entry, as summarized in Algorithm~\ref{alg:tram_route}.
Let $\mathbf{T}^{\ell}\in\mathbb{R}^{B\times L\times C}$ and $\mathbf{H}^{\ell}\in\mathbb{R}^{B\times N\times C}$ be text and image tokens at layer $\ell$.
Token compaction is applied strictly to image tokens $\mathbf{H}^{\ell}$, while text tokens remain entirely untouched.
Temporal curvature is used only as a lightweight saliency cue for token ranking under a fixed routing budget; it does not modify the denoising trajectory.

\textbf{Text-similarity importance.}
Let $\bar{\mathbf{t}}_b=\frac{1}{L}\sum_{j=1}^{L}\mathbf{T}^{\ell}_{b,j}$ be the mean text embedding.
For token $n$:
\begin{equation}
s^{\mathrm{text}}_{b,n}
=
\mathrm{Norm}\!\left(
\frac{\langle \mathbf{H}^{\ell}_{b,n},\bar{\mathbf{t}}_b\rangle}
{\|\mathbf{H}^{\ell}_{b,n}\|_2\|\bar{\mathbf{t}}_b\|_2+\epsilon}
\right),
\label{eq:text_score}
\end{equation}
where $\mathrm{Norm}(\cdot)$ denotes per-sample min--max normalization over image tokens.

\textbf{Temporal Curvature Guidance (TCG) importance.}
We reshape image tokens as $\mathbf{X}\in\mathbb{R}^{B\times F_g\times P\times C}$, where $F_g$ is the number of frame groups and $P$ is the number of tokens per group.
Define temporal velocity and curvature:
\begin{equation}
\mathbf{v}_{b,f,p}=\mathbf{X}_{b,f+1,p}-\mathbf{X}_{b,f,p},
\label{eq:tcg_vel}
\end{equation}
\begin{equation}
\kappa_{b,f,p}
=
\arccos\!\left(
\frac{\langle \mathbf{v}_{b,f,p},\mathbf{v}_{b,f+1,p}\rangle}
{\|\mathbf{v}_{b,f,p}\|_2\|\mathbf{v}_{b,f+1,p}\|_2+\epsilon}
\right),
\label{eq:tcg_curv}
\end{equation}
where the cosine term is clamped to $[-1,1]$ prior to the $\arccos$ operation for numerical stability.
We reshape/broadcast $\kappa$ back to the flattened token order ($N=F_gP$) and normalize to obtain $s^{\mathrm{tcg}}\in\mathbb{R}^{B\times N}$.

\textbf{Semantic--temporal importance and keep set.}
We fuse cues by normalized weighted averaging:
\begin{equation}
s_{b,n}
=
\mathrm{Norm}\!\left(
\frac{w_t\,s^{\mathrm{text}}_{b,n}+w_p\,s^{\mathrm{tcg}}_{b,n}}
{w_t+w_p}
\right).
\label{eq:score_fusion}
\end{equation}
Given the fixed drop ratio $\rho$, the keep count is $K_r=\max(1,N-\lfloor \rho N\rfloor)$.
We select the keep/drop sets by top-$K_r$:
\begin{equation}
\mathcal{K}_r=\mathrm{TopK}(s,K_r),\qquad
\mathcal{D}_r=\{1,\dots,N\}\setminus\mathcal{K}_r.
\label{eq:keep_drop}
\end{equation}

\subsection{Two-Stream Token Merge and Identity Restoration}
\label{sec:two_stream}

Inside a routed interval, computation proceeds via a two-stream architecture (see Algorithm~\ref{alg:tram_route}): a local stream for kept tokens and a compact global stream formed by merging dropped tokens.

\textbf{Route entry (merge).}
Kept tokens are gathered as
\begin{equation}
\mathbf{H}_{\mathrm{loc}}=\mathrm{Gather}(\mathbf{H}^{\ell},\mathcal{K}_r).
\label{eq:gather_loc}
\end{equation}
Let $M_r=|\mathcal{D}_r|$ and global ratio $\gamma$.
The number of global tokens is bounded by
\begin{equation}
G_r=\min\!\Big(M_r,\ g_{\max},\ \max(g_{\min},\lfloor \gamma M_r\rfloor)\Big).
\label{eq:global_count}
\end{equation}
We sort $\mathcal{D}_r$ and partition it into $G_r$ groups $\{\mathcal{D}_{r,g}\}_{g=1}^{G_r}$.
Each group is merged by mean pooling:
\begin{equation}
\mathbf{g}_{b,g}
=
\frac{1}{|\mathcal{D}_{r,g}|}
\sum_{n\in\mathcal{D}_{r,g}}
\mathbf{H}^{\ell}_{b,n}.
\label{eq:mean_merge}
\end{equation}
The working tokens are
\begin{equation}
\mathbf{H}^{\ell}_{\mathrm{work}}
=
[\mathbf{H}_{\mathrm{loc}};\mathbf{G}],\quad
\mathbf{G}=[\mathbf{g}_{1},\dots,\mathbf{g}_{G_r}].
\label{eq:work_tokens}
\end{equation}
For rotary embeddings, local tokens use routed indices, while each global token uses averaged rotary features within its group.

\textbf{Route exit (unmerge).} Let $\mathbf{H}_{\mathrm{work}}^{\ell'}=[\mathbf{H}_{\mathrm{loc}}^{\ell'};\mathbf{G}^{\ell'}]$ be the output at the end layer $e_r$. To reconstruct the full-resolution token grid, we bypass the global stream features and populate the dropped positions with their original features saved at the route entry (denoted as $\mathbf{H}_{\mathrm{snap}}$):
\begin{equation}
\tilde{\mathbf{H}}_{b,n}^{\ell'}
=
\begin{cases}
\mathbf{H}_{\mathrm{loc},b,\pi(n)}^{\ell'}, & n\in\mathcal{K}_r,\\
\mathbf{H}_{\mathrm{snap},b,n}, & n\in\mathcal{D}_r,
\end{cases}
\label{eq:restore_snapshot}
\end{equation}
where $\pi(\cdot)$ maps an original kept index to its local position.
This mechanism reduces attention complexity from $\mathcal{O}(N^2)$ to approximately $\mathcal{O}((K_r+G_r)^2)$ inside routed layers.

\textbf{Rationale.}
The local stream preserves high-fidelity tokens for detail synthesis, while the global stream retains low-cost context within routed layers.
Identity restoration avoids injecting potentially noisy updates into dropped positions and yields stable behavior under aggressive routing.

\begin{algorithm}[t]
\caption{TRaM-VSR Route-and-Merge in a Routed Interval}
\label{alg:tram_route}
\small
\begin{algorithmic}[1]
\Require Full tokens $X\in\mathbb{R}^{N\times C}$, text tokens $T$, drop ratio $\rho$, global ratio $\gamma$, weights $(w_t,w_p)$, routed blocks $\mathcal{F}$
\Ensure Restored full tokens $\hat{X}\in\mathbb{R}^{N\times C}$
\State \textbf{Semantic--temporal scoring:}
\State \quad $s^{text} \gets \mathrm{TextSim}(X, T)$
\State \quad $s^{tcg} \gets \mathrm{TCG}(X)$ \Comment{time-group curvature cue}
\State \quad $s \gets \mathrm{Norm}(w_t s^{text} + w_p s^{tcg})$
\State \textbf{Top-$K$ keep:}
\State \quad $K \gets \max(1, N - \lfloor \rho N \rfloor)$
\State \quad $\mathcal{K}_r \gets \mathrm{TopK}(s, K)$;\; $\mathcal{D}_r \gets [1..N]\setminus \mathcal{K}_r$
\State \textbf{Two-stream merge:}
\State \quad $X_{loc} \gets X[\mathcal{K}_r]$
\State \quad $X_{glob} \gets \mathrm{GroupMean}(X[\mathcal{D}_r], \gamma)$
\State \quad $X_{work} \gets [X_{loc}; X_{glob}]$
\State \textbf{Routed computation:} $Y_{work} \gets \mathcal{F}(X_{work})$
\State \textbf{Unmerge:}
\State \quad $\hat{X} \gets X$ \Comment{freeze base for dropped tokens}
\State \quad $\hat{X}[\mathcal{K}_r] \gets Y_{work}[1{:}K]$ \Comment{scatter kept tokens}
\State \Return $\hat{X}$
\end{algorithmic}
\end{algorithm}
\section{Experiments}
\label{sec:experiments}
\begin{figure}[t]
  \centering
  \includegraphics[width=\linewidth]{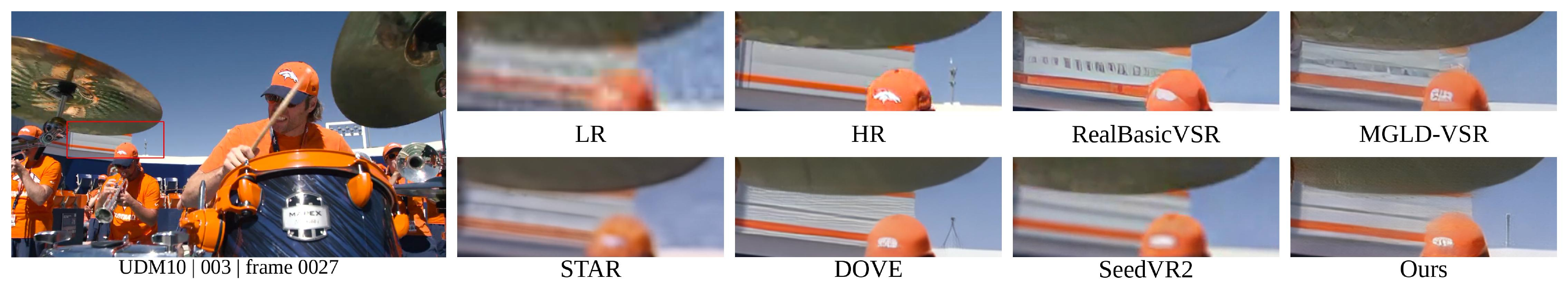}
  \includegraphics[width=\linewidth]{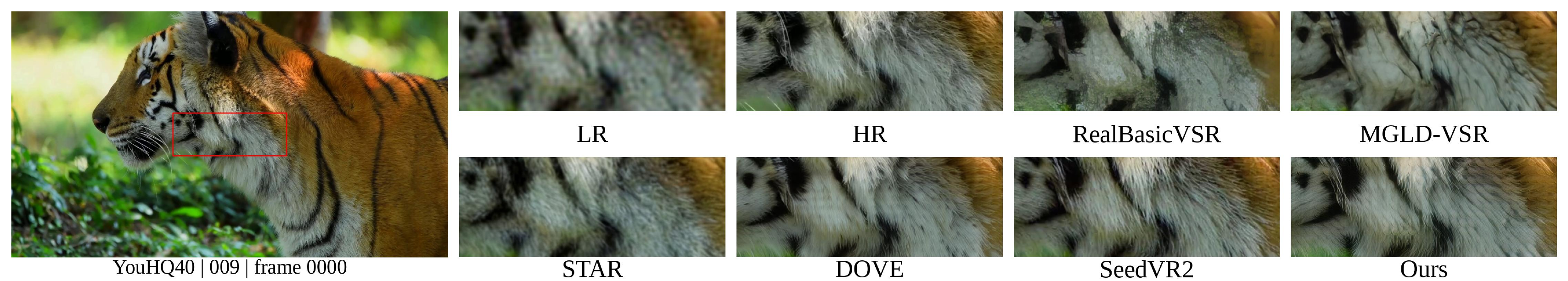}
  \caption{
  Visual comparison on real-world UDM10~\cite{tao2017detail} (top) and synthetic YouHQ40~\cite{zhou2024upscale} (bottom). TRaM-VSR recovers substantially sharper fine textures and structurally coherent edges compared to recent baselines.
  }
  \label{fig:visual_comp}
\end{figure}

\subsection{Experimental Settings}
\label{sec:exp_settings}

\noindent \textbf{Datasets.}
Following standard VSR evaluation protocols, all experiments are conducted at a $\times 4$ upscaling factor. We evaluate our method on three synthetic benchmarks (UDM10~\cite{tao2017detail}, SPMCS~\cite{yi2019progressive}, and YouHQ40~\cite{zhou2024upscale}) and three real-world benchmarks (RealVSR~\cite{yang2021real}, MVSR4x~\cite{wang2023benchmark}, and VideoLQ~\cite{chan2022investigating}). The synthetic datasets employ the same degradation pipeline used during training. RealVSR and MVSR4x contain real-world LQ--HQ pairs captured in the wild, whereas VideoLQ consists of internet-sourced videos without ground-truth HR.

\noindent \textbf{Evaluation Metrics.}
For datasets with HR references, we report standard fidelity metrics (PSNR and SSIM~\cite{wang2004image}) and perceptual full-reference metrics (LPIPS~\cite{zhang2018unreasonable} and DISTS~\cite{ding2020image}). For no-reference perceptual evaluation, we employ CLIP-IQA~\cite{wang2023exploring} and MUSIQ~\cite{ke2021musiq}. To strictly assess temporal consistency, we measure the flow-warping error ($E^*_{warp}$)~\cite{lai2018learning}, where lower values indicate smoother temporal transitions. Unless otherwise stated, all metrics are computed in the RGB color space.

\noindent \textbf{Implementation Details.}
TRaM-VSR is built upon a one-step latent DiT VSR architecture and is evaluated using a fixed one-step sampling scheme. To ensure fair evaluation, we maintain strictly identical degradation and testing protocols across all compared methods. By default, TRaM-VSR is configured with two routed intervals, an importance fusion mechanism combining text similarity and TCG cues, and the two-stream local--global token architecture with identity restoration at the route exits. At inference time, TRaM-VSR does not require any manually provided caption or prompt; all quantitative and visual results are obtained under this prompt-free setting.

\subsection{Comparison with State-of-the-Art Methods}
\label{sec:exp_sota}

We compare TRaM-VSR against representative state-of-the-art image and video restoration baselines, including RealESRGAN~\cite{wang2021real}, ResShift~\cite{yue2025efficient}, RealBasicVSR~\cite{chan2022investigating}, Upscale-A-Video~\cite{zhou2024upscale}, MGLD-VSR~\cite{yang2024motion}, STAR~\cite{xie2025star}, and DOVE~\cite{chen2025dove}. 
\begin{table*}[t]
\centering
\setlength{\tabcolsep}{3.0pt}
\renewcommand{\arraystretch}{1.12}
\caption{Quantitative comparison with state-of-the-art methods. Best and second-best results are marked in \textcolor{red}{red} and \textcolor{blue}{blue}.}
\resizebox{\linewidth}{!}{%
\begin{tabular}{llccccccccc}
\toprule

\shortstack[c]{Dataset\\\vphantom{\cite{wang2021real}}}
& \shortstack[c]{Metric\\\vphantom{\cite{wang2021real}}}
& \shortstack[c]{Real- \\ ESRGAN\cite{wang2021real}}
& \shortstack[c]{ResShift\\\cite{yue2025efficient}}
& \shortstack[c]{RealBasicVSR\\\cite{chan2022investigating}}
& \shortstack[c]{Upscale-A- \\ Video\cite{zhou2024upscale}}
& \shortstack[c]{MGLD- \\ VSR\cite{yang2024motion}}
& \shortstack[c]{STAR\\\cite{xie2025star}}
& \shortstack[c]{DOVE\\\cite{chen2025dove}}
& \shortstack[c]{SeedVR2- \\ 7B\cite{wang2025seedvr2}}
& \shortstack[c]{Ours\\\vphantom{\cite{wang2021real}}} \\

\midrule

\multirow{6}{*}{UDM10}
& PSNR $\uparrow$
& 24.04 & 23.65 & 24.13 & 21.72 & 24.23 & 23.47 & \textcolor{blue}{26.48} & 25.47 & \textcolor{red}{26.73} \\
& SSIM $\uparrow$
& 0.7107 & 0.6016 & 0.6801 & 0.5913 & 0.6957 & 0.6804 & \textcolor{red}{0.7827} & 0.7582 & \textcolor{blue}{0.7759} \\
& LPIPS $\downarrow$
& 0.3877 & 0.5537 & 0.3908 & 0.4116 & 0.3272 & 0.4242 & 0.2696 & \textcolor{red}{0.2507} & \textcolor{blue}{0.2678} \\
& DISTS $\downarrow$
& 0.2184 & 0.2898 & 0.2067 & 0.2230 & 0.1677 & 0.2156 & 0.1492 & \textcolor{red}{0.1263} & \textcolor{blue}{0.1450} \\
& CLIP-IQA $\uparrow$
& 0.4189 & 0.4344 & 0.3494 & \textcolor{blue}{0.4697} & 0.4557 & 0.2417 & \textcolor{red}{0.5107} & 0.2958 & 0.4547 \\
& $E^*_{warp}\downarrow$
& 4.83 & 6.12 & 3.10 & 3.97 & 3.59 & 2.08 & \textcolor{blue}{1.77} & 2.20 & \textcolor{red}{1.73} \\
\midrule

\multirow{6}{*}{SPMCS}
& PSNR $\uparrow$
& 21.22 & 21.68 & 22.17 & 18.81 & 22.39 & 21.24 & \textcolor{blue}{23.11} & 22.53 & \textcolor{red}{23.18} \\
& SSIM $\uparrow$
& 0.5613 & 0.5153 & 0.5638 & 0.4113 & 0.5896 & 0.5441 & \textcolor{blue}{0.6210} & 0.6187 & \textcolor{red}{0.6237} \\
& LPIPS $\downarrow$
& 0.3721 & 0.4467 & 0.3662 & 0.4468 & 0.3263 & 0.5257 & 0.2888 & \textcolor{red}{0.2781} & \textcolor{blue}{0.2883} \\
& DISTS $\downarrow$
& 0.2220 & 0.2697 & 0.2164 & 0.2452 & 0.1960 & 0.2872 & 0.1713 & \textcolor{red}{0.1530} & \textcolor{blue}{0.1613} \\
& CLIP-IQA $\uparrow$
& 0.5238 & 0.5442 & 0.3513 & 0.5248 & 0.4348 & 0.2646 & \textcolor{red}{0.5690} & 0.4271 & \textcolor{blue}{0.5588} \\
& $E^*_{warp}\downarrow$
& 5.61 & 8.07 & 1.88 & 4.22 & 1.68 & \textcolor{blue}{1.01} & 1.04 & 1.38 & \textcolor{red}{0.97} \\
\midrule

\multirow{6}{*}{YouHQ40}
& PSNR $\uparrow$
& 22.82 & 23.32 & 22.39 & 19.62 & 23.17 & 22.64 & \textcolor{blue}{24.30} & 23.30 & \textcolor{red}{24.38} \\
& SSIM $\uparrow$
& 0.6337 & 0.6273 & 0.5895 & 0.4824 & 0.6194 & 0.6323 & \textcolor{blue}{0.6740} & 0.6659 & \textcolor{red}{0.6856} \\
& LPIPS $\downarrow$
& 0.3571 & 0.4211 & 0.4091 & 0.4268 & 0.3608 & 0.4600 & 0.2997 & \textcolor{red}{0.2705} & \textcolor{blue}{0.2903} \\
& DISTS $\downarrow$
& 0.1790 & 0.2159 & 0.1933 & 0.2012 & 0.1685 & 0.2287 & 0.1477 & \textcolor{red}{0.1117} & \textcolor{blue}{0.1370} \\
& CLIP-IQA $\uparrow$
& 0.4704 & 0.4633 & 0.3964 & \textcolor{red}{0.5258} & 0.4657 & 0.2739 & \textcolor{blue}{0.4985} & 0.3321 & 0.4676 \\
& $E^*_{warp}\downarrow$
& 5.91 & 5.75 & 3.08 & 6.84 & 3.45 & 2.21 & \textcolor{blue}{2.05} & 4.03 & \textcolor{red}{1.70} \\
\midrule

\multirow{6}{*}{RealVSR}
& PSNR $\uparrow$
& 20.85 & 20.81 & 22.12 & 20.29 & 22.02 & 17.43 & \textcolor{blue}{22.32} & 22.09 & \textcolor{red}{22.37} \\
& SSIM $\uparrow$
& 0.7105 & 0.6277 & 0.7163 & 0.5945 & 0.6774 & 0.5215 & \textcolor{red}{0.7301} & 0.7181 & \textcolor{blue}{0.7288} \\
& LPIPS $\downarrow$
& 0.2016 & 0.2312 & \textcolor{blue}{0.1870} & 0.2671 & 0.2182 & 0.2943 & \textcolor{red}{0.1851} & 0.2014 & 0.1911 \\
& DISTS $\downarrow$
& 0.1279 & 0.1435 & \textcolor{blue}{0.0983} & 0.1425 & 0.1169 & 0.1599 & \textcolor{red}{0.0978} & 0.1146 & 0.1005 \\
& CLIP-IQA $\uparrow$
& \textcolor{red}{0.7472} & \textcolor{blue}{0.5553} & 0.2905 & 0.4855 & 0.4510 & 0.3641 & 0.5207 & 0.2872 & 0.5062 \\
& $E^*_{warp}\downarrow$
& 6.32 & 9.55 & 4.45 & 6.25 & \textcolor{red}{3.16} & 9.88 & 3.52 & 3.54 & \textcolor{blue}{3.23} \\

\midrule
\multirow{6}{*}{MVSR4x}
& PSNR $\uparrow$
& 22.47 & 21.58 & 21.80 & 20.42 & \textcolor{red}{22.77} & 22.42 & 22.42 & 22.46 & \textcolor{blue}{22.56} \\
& SSIM $\uparrow$
& 0.7412 & 0.6473 & 0.7045 & 0.6117 & 0.7418 & 0.7421 & \textcolor{blue}{0.7523} & 0.7426 & \textcolor{red}{0.7591} \\
& LPIPS $\downarrow$
& 0.4534 & 0.5945 & 0.4235 & 0.4717 & 0.3568 & 0.4311 & \textcolor{blue}{0.3476} & 0.3643 & \textcolor{red}{0.3373} \\
& DISTS $\downarrow$
& 0.3021 & 0.3351 & 0.2498 & 0.2673 & \textcolor{red}{0.2245} & 0.2714 & 0.2363 & \textcolor{blue}{0.2293} & \textcolor{red}{0.2245} \\
& CLIP-IQA $\uparrow$
& 0.4396 & 0.5003 & 0.4118 & \textcolor{red}{0.6106} & 0.3769 & 0.2674 & \textcolor{blue}{0.5453} & 0.2127 & 0.5076 \\
& $E^*_{warp}\downarrow$
& 1.64 & 3.89 & 1.69 & 5.10 & 1.55 & \textcolor{red}{0.61} & 0.78 & 0.92 & \textcolor{blue}{0.75} \\
\bottomrule
\end{tabular}%
} 

\label{tab:quantitative}
\end{table*}
\noindent \textbf{Quantitative Results.}
Quantitative comparisons on synthetic and real-world benchmarks are reported in Table~\ref{tab:quantitative}. Overall, TRaM-VSR achieves a superior quality--efficiency balance under strictly one-step inference. On synthetic datasets, our framework consistently improves perceptual metrics (e.g., LPIPS and DISTS) while maintaining highly competitive pixel-wise fidelity (PSNR/SSIM). Crucially, on the RealVSR dataset, TRaM-VSR remains robust, yielding improved perceptual scores and strong temporal consistency. These results substantiate the effectiveness of our importance-aware routing and two-stream token processing for precision-critical video restoration.

\noindent \textbf{Qualitative Results.}
Visual comparisons on real-world UDM10 and synthetic YouHQ40 are provided in Figure~\ref{fig:visual_comp}. TRaM-VSR synthesizes sharper structures and more faithful high-frequency textures, effectively avoiding the over-smoothing artifacts typical of purely regression-based restorers or aggressive token compaction. In particular, our method reliably preserves repetitive patterns and thin structural edges.

\begin{figure}[t]
  \centering
  \includegraphics[width=\linewidth]{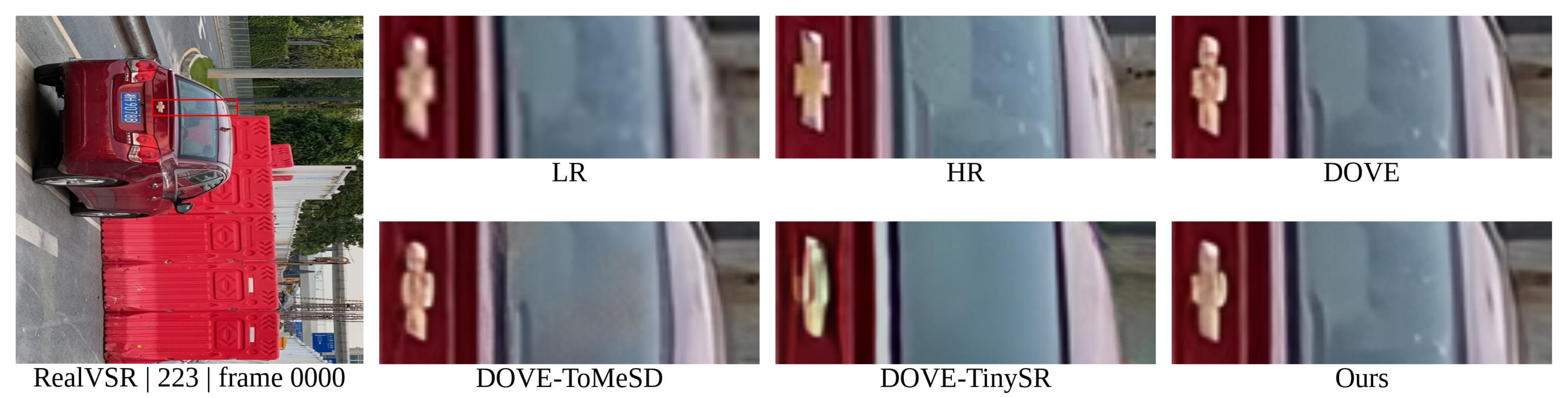}
  \includegraphics[width=\linewidth]{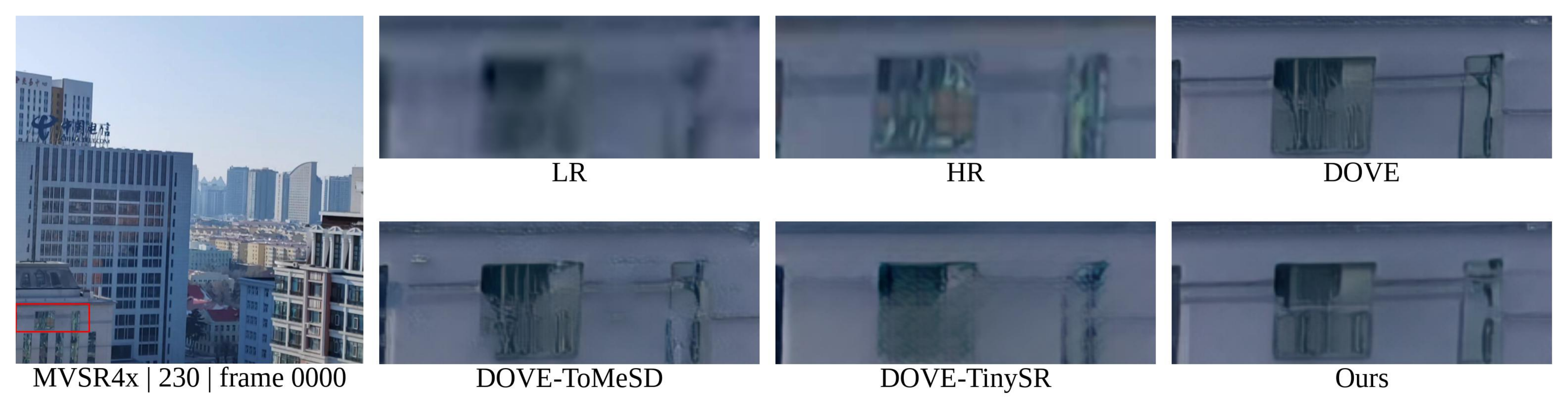}
  \caption{
Visual comparison of alternative acceleration strategies applied to the one-step DOVE backbone on RealVSR~\cite{yang2021real} (top) and MVSR4x~\cite{wang2023benchmark} (bottom). Compared to uniform token merging (DOVE-ToMeSD) and token pruning (DOVE-TinySR), our TRaM-VSR effectively mitigates texture over-smoothing and strictly preserves structural integrity under high compression regimes.
  }
  \label{fig:visual_comp_real}
\end{figure}

\noindent \textbf{Efficacy of the Routing Strategy.}
To explicitly isolate the benefits of our framework, we compare TRaM-VSR against alternative acceleration strategies applied to the identical one-step DiT backbone: (i) uniform token merging and (ii) uniform token pruning (Figure~\ref{fig:visual_comp_real}). While naive merging and pruning reduce computational overhead, they consistently introduce characteristic artifacts such as texture over-smoothing, local contrast drift, and inconsistent edge transitions—flaws that are exacerbated by the skip-free nature of DiT architectures. In contrast, TRaM-VSR leverages semantic--temporal scoring to safeguard visually vital regions and preserves global context via the two-stream representation. Furthermore, identity restoration definitively prevents the irreversible detail loss observed in the baselines. Consequently, our method reconstructs sharper, more coherent textures, demonstrating a strictly more favorable quality--efficiency trade-off for real-world scenarios.

\begin{figure}[t]
  \centering
  \includegraphics[width=\linewidth]{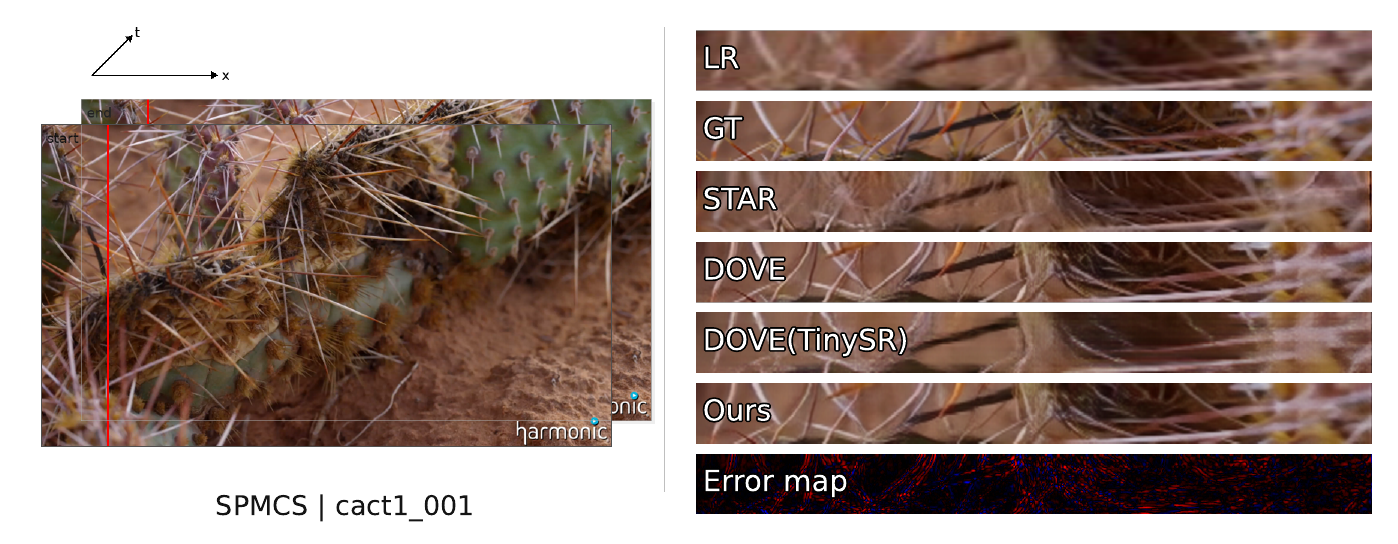}
  \caption{
  Temporal consistency visualization on SPMCS by stacking the red vertical line across consecutive frames (left) and showing the corresponding temporal slices (right); the last row is the error map (red indicates larger error).
  }
  \label{fig:temporal_consistency}
\end{figure}

\noindent \textbf{Temporal Consistency.}
We visualize temporal coherence using space--time slice inspection in Figure~\ref{fig:temporal_consistency}. While prior baselines exhibit noticeable structural wobbling and temporal fluctuations under complex degradations (reflected by significant slice errors), TRaM-VSR yields remarkably smooth temporal profiles with minimized error accumulation. This visual stability strongly aligns with the superior $E^*_{warp}$ performance reported in Table~\ref{tab:quantitative}.

\begin{figure}[t]
    \centering
    \includegraphics[width=\linewidth]{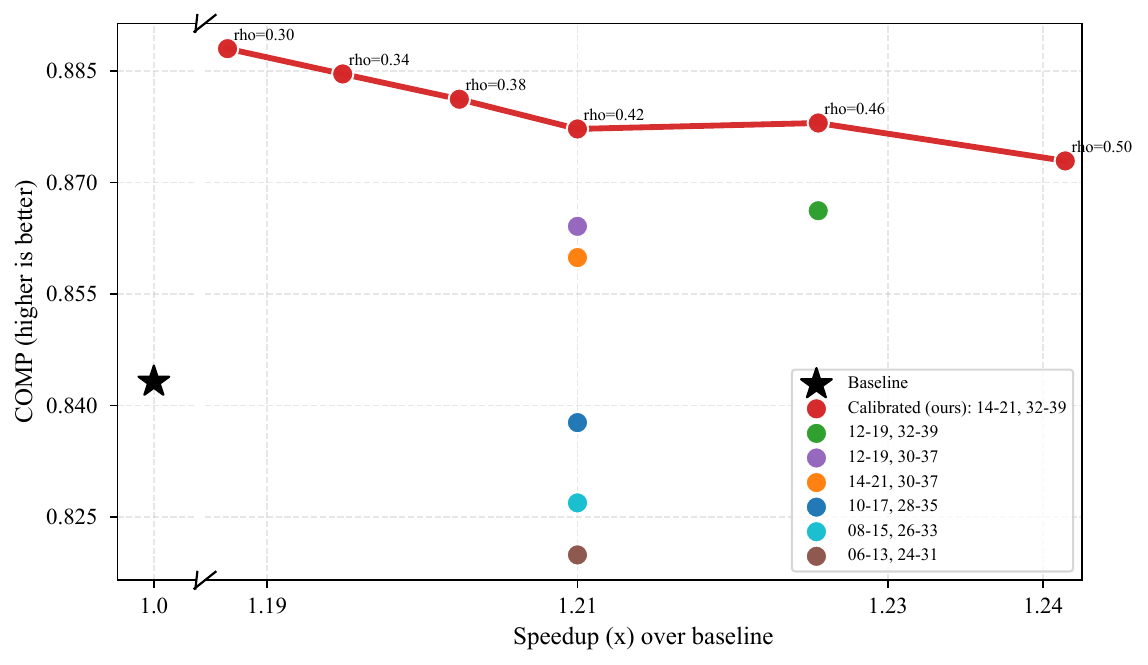}
    \caption{\textbf{Speed--quality trade-off under different routed intervals and drop ratios. }
    }
    \label{fig:route_tradeoff}
\end{figure}

\definecolor{OursRow}{RGB}{230,245,255}




\subsection{Ablation Study}
\label{sec:exp_ablation}

We conduct extensive ablations on the SPMCS dataset to validate the core architectural designs of TRaM-VSR.

\noindent \textbf{Routing Structure.}
Table~\ref{tab:ablation_struct_spmcs} compares random routing, single-stream importance-aware routing, and our proposed two-stream framework. Random routing severely degrades perceptual quality due to unstructured token removal. While implementing importance-aware routing improves selection, the single-stream variant (lacking a global stream) still suffers perceptually. This indicates that aggressively discarding tokens without retaining global context is profoundly detrimental to VSR. Our full two-stream design achieves the best overall perceptual quality (LPIPS and MUSIQ) while maintaining robust temporal consistency, demonstrating the necessity of jointly modeling localized high-fidelity details and compact global context within routed layers.

\noindent \textbf{Importance Fusion.}
We ablate the components of our token-scoring mechanism in Table~\ref{tab:ablation_ratio_spmcs_single}. Relying solely on text similarity provides semantic guidance but neglects motion-sensitive regions. Conversely, utilizing TCG scoring alone strictly enforces temporal dynamics (yielding the lowest $E^*_{warp}$ error) but limits semantic synthesis. Fusing text similarity and TCG cues (at a 0.6/0.4 ratio) achieves the optimal balance, maximizing overall perceptual quality without severely compromising video-level fidelity. This confirms that semantic relevance and temporal dynamics offer complementary signals for identifying precision-critical tokens.

\noindent \textbf{Offline Planner and Route Scheduling.}
The efficacy of our routed-interval selection is evaluated via the speed--quality analysis in Figure~\ref{fig:route_tradeoff}. Each plotted point represents a distinct interval configuration and drop ratio. The specific routing intervals identified by our offline planner consistently trace a superior quality--efficiency Pareto front compared to alternative heuristic placements at comparable speedups. This validates our calibration-guided scheduling approach. Furthermore, sweeping the drop ratio produces a smooth trade-off curve, confirming that $\rho$ serves as a predictable and robust control knob for practical deployment.

\noindent \textbf{Identity Restoration.}
Finally, we analyze the critical role of route-exit restoration. In standard skip-free DiT backbones, intermediate token compaction inevitably induces irreversible detail loss. By completely bypassing the global stream and reconstructing the full-resolution grid using untouched features saved at the interval entry, our identity restoration mechanism drastically improves structural robustness. This ensures that precision-critical details remain pristine while fully retaining the computational benefits of the compacted routed layers (as evidenced by Table~\ref{tab:ablation_struct_spmcs}).
\begin{figure}[t]
  \centering
  \includegraphics[width=\linewidth]{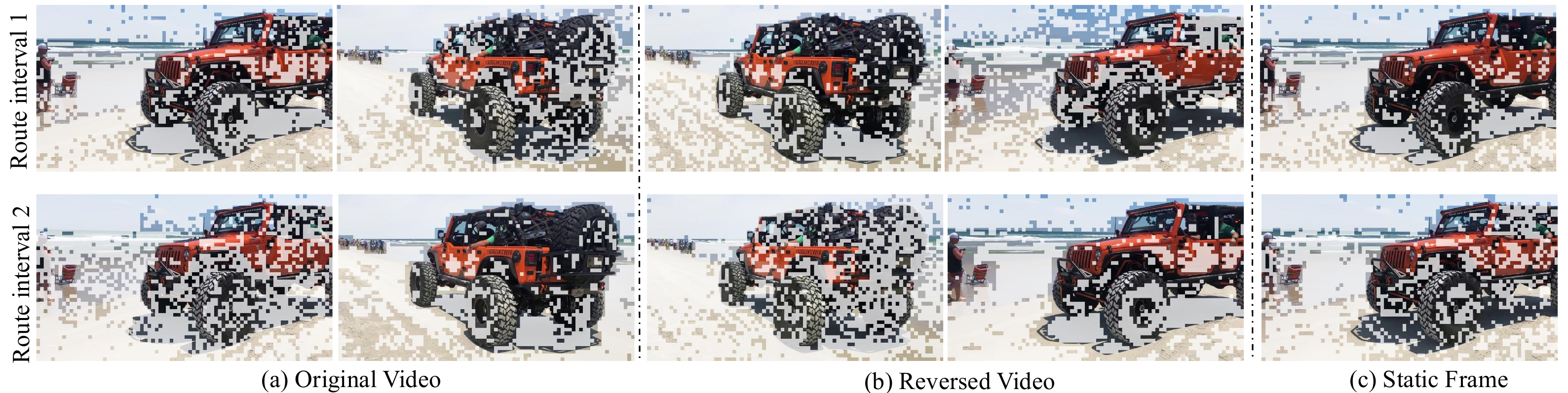}
  \caption{Spatio-temporal routing visualization on the dynamic ``Jeep'' sequence.}
  \label{fig:motivation}
\end{figure}

\begin{table}[t]
\scriptsize
\centering
\setlength{\tabcolsep}{3.0pt}
\renewcommand{\arraystretch}{1.08}
\caption{Structural ablation on SPMCS: random routing vs.\ single-stream routing vs.\ two-stream local/global routing. Best results are marked in \textbf{bold}.
}
\begin{tabular}{lcccc}
\toprule
Variant & PSNR $\uparrow$ & LPIPS $\downarrow$ & MUSIQ $\uparrow$ & $E^*_{warp}\downarrow$ \\
\midrule
Random route (single-stream)            & 23.13 & 0.3187 & 62.84 & 1.0030 \\
Text+TCG (single-stream, no global)     & 23.08 & 0.3391 & 63.28 & 0.9839 \\
Full TRaM-VSR (two-stream, with global) & \textbf{23.18} & \textbf{0.2883} & \textbf{67.27} &  \textbf{0.9720} \\
\bottomrule
\end{tabular}

\label{tab:ablation_struct_spmcs}
\end{table}

\begin{table}[t]
\scriptsize
\centering
\setlength{\tabcolsep}{3.0pt}
\renewcommand{\arraystretch}{1.08}
\caption{Ablation on SPMCS. The 0.6/0.4 fusion setting yields the best overall perceptual quality, while utilizing TCG-only achieves the lowest warp error. Best results are marked in \textbf{bold}.
}
\begin{tabular}{lcccc}
\toprule
Variant (SPMCS, inference) & PSNR $\uparrow$ & LPIPS $\downarrow$ & MUSIQ $\uparrow$ & $E^*_{warp}\downarrow$ \\
\midrule
Text-only          & 23.09 & 0.3450 & 62.79 & 0.9295 \\
PSG-only           & 22.98 & 0.3343 & 59.49 & \textbf{0.9068} \\
Text+PSG (0.6/0.4) & \textbf{23.18} & \textbf{0.2883} & \textbf{67.27} & 0.9720 \\
\bottomrule
\end{tabular}

\label{tab:ablation_ratio_spmcs_single}
\end{table}

\section{Token Visualization and Discussion}
\label{discussion}

In Figure \ref{fig:motivation}, we visualize the token selection for the video examples shown in Figure \ref{fig:introduction}. From left to right, we present: (a) token selection for the original video, (b) for the reversed video, and (c) for a static video where the first frame is repeated across all video frames, resulting in no changes.

\noindent \textbf{Asymmetric Token Routing.} Examining case (a) in detail, we observe that routing decisions differ between block intervals 1 and 2. More importantly, when the car is approaching (left), tokens corresponding to the car’s shadow are dropped more aggressively, whereas when the car is leaving (right), the trend is reversed. We hypothesize that this behavior reflects the natural motion dynamics and the model’s implicit world understanding of the car’s movement. As the car approaches, its front shadow gradually expands. Although this introduces local pixel changes, the newly appearing shadow regions follow a highly regular and predictable pattern, progressively turning local areas into darker pixels and reducing their informative content. As a result, our method identifies these regions as statistically predictable and applies more aggressive dropping.

Interestingly, compared to shallow block segments, deeper blocks exhibit slightly higher activation on the same region. While this may appear to introduce additional computation, it actually reflects a meaningful hierarchical modeling strategy, which is further facilitated by our network-level prior. Although the shadow itself is predictable at the pixel level, it remains closely related to the object’s geometric structure, illumination relationships, and temporal continuity. Consequently, the network-level prior encourages deeper layers, which operate at a more semantic and structural level, to retain limited attention on these regions in order to maintain geometric and temporal consistency between the shadow and the moving object.

\noindent \textbf{Emerging Bidirectional Temporal Behavior.} We further analyze the reversed video and visualize the same frames reordered as in (b). In this case, the car moves backward and the shadow continues to expand. Here we observe a similar activation trend as in (a). This observation is particularly notable because the baseline model~\cite{chen2025dove} was originally trained only for single-direction temporal processing. However, after fine-tuning the baseline with our proposed network, our token adaptation mechanism enables the originally unidirectional model to effectively handle bidirectional temporal dynamics in a consistent way.

\noindent \textbf{Without Motion.} Finally, we manually construct a static video by repeating the first frame, as illustrated in (c), such that no motion exists throughout the sequence. This setting allows us to isolate the behavior of the network-level priors. Interestingly, when comparing (a-left) with (c), we observe that the deeper block segments exhibit substantially higher activation than the shallow ones in the same shadow regions. As motion cues vanish in this static setting, shadows are no longer predictable from temporal dynamics.  Instead, we venture that in this case shadows become stable spatial structures tightly coupled with object geometry and illumination, carrying greater semantic significance. Consequently, deeper blocks tend to preserve these regions to maintain semantic and geometric consistency, leading to reduced dropping compared to the dynamic scenario.
\section{Limitations and Future Work}
Although TRaM-VSR improves temporal stability and efficiency, several limitations remain. First, one-step VSR may still show mild residual oscillation in challenging regions such as fine repetitive textures, shadows, and fast local motion. Second, our speedup primarily targets DiT token computation, so the end-to-end acceleration depends on the relative cost of VAE decoding and implementation details such as tiling. Third, the routed intervals are selected by offline calibration; while this setup is fixed across our benchmarks, new backbones or substantially different degradation regimes may benefit from recalibration. Future work will explore stronger temporal priors and adaptive routing calibration.
\section{Conclusion}
\label{sec:conclusion}

In this paper, we presented TRaM-VSR, an efficient Token Routing and Merging framework that addresses the prohibitive computational costs and temporal instability inherent in one-step DiT-based video super-resolution. Moving beyond uniform token compression, our approach dynamically allocates computational budgets by jointly leveraging context-aware video priors and multigranular network-level priors. By fusing semantic-temporal importance scoring with depth-calibrated routing, TRaM-VSR explicitly preserves dynamic objects and structural boundaries within a high-fidelity local stream, while compactly aggregating redundant tokens. Extensive evaluations confirm that our framework significantly accelerates inference without compromising state-of-the-art perceptual quality and temporal coherence, offering a highly practical solution for real-world VSR deployment.

\paragraph{Acknowledgments.}
This work was supported by The Alexander von Humboldt Foundation.

{\small
\bibliographystyle{splncs04}
\bibliography{main}
}

\end{document}